\documentclass[runningheads]{llncs}
\usepackage[T1]{fontenc}

\def\paperstate{submit} 

\newif\ifdraft
\newif\ifsubmit

\def\stateD{draft}
\def\stateS{submit}

\ifx\paperstate\stateD
  \drafttrue   \submitfalse
\else\ifx\paperstate\stateS
  \draftfalse  \submittrue
\else
  \errmessage{Invalid paperstate}
\fi\fi

\usepackage{graphicx}
\usepackage{bm}
\usepackage{cite}
\usepackage{amsmath}  
\usepackage{amssymb}   
\usepackage{booktabs}
\usepackage{subcaption}
\usepackage{siunitx}
\usepackage{enumitem}
\usepackage{multirow}
\usepackage[font=normal,labelfont=bf]{caption}
\usepackage{cases}
\usepackage{pgfplots}
    \pgfplotsset{compat=1.18}
\usepackage[ruled, vlined]{algorithm2e}
\usepackage{svg}
\usepackage{tabularx} 

\newcommand{\LO}{X^\ast}
\newcommand{\sol}[1]{\bm{#1}}

\usepackage{cleveref}
    \crefname{figure}{Fig.}{Figs.}
    \Crefname{figure}{Fig.}{Figs.}
    \crefname{table}{Table}{Tables}
    \crefname{algorithm}{Algorithm}{Algorithms}
    \crefname{equation}{Eq.}{Eqs.}
    \crefname{section}{Section}{Sections}
    \crefname{subfigure}{Fig.}{Figs.}
    \Crefname{subfigure}{Fig.}{Figs.}
    \crefformat{subfigure}{#2(#1)#3}

\ifdraft
  \usepackage{xcolor}
  \newcommand{\todo}[1]{{\textcolor{red}{\bfseries [TODO: #1]}}}
  
\else
  \newcommand{\todo}[1]{}
  
\fi
\newcommand{\hide}[1]{}
\begin{document}

\title{Novelty-Based Generation of Continuous Landscapes with Diverse Local Optima Networks}
\titlerunning{Novelty-Based Generation of Continuous Landscapes with Diverse LONs}
%
%
\author{Kippei~Mizuta\inst{}$^{\dagger}$\orcidID{0009-0004-5332-2131}\and
Shoichiro~Tanaka\inst{}$^{\dagger}$\orcidID{0009-0002-9230-445X}\and
Shuhei~Tanaka\orcidID{0009-0006-3807-4437}\and
Toshiharu~Hatanaka\orcidID{0000-0003-3670-6825}}
\authorrunning{K. Mizuta et al.}
\institute{The University of Fukuchiyama, Fukuchiyama, Kyoto, Japan\\
\email{\{tanaka-shoichiro, hatanaka-toshiharu\}@fukuchiyama.ac.jp}}
\maketitle
\begin{abstract}
\textit{Local Optima Networks} (LONs) encode the global structure of search spaces as graphs, but their construction requires iterative execution of a search algorithm to identify local optima and approximate the transitions between \textit{Basins of Attraction} (BoAs).
In continuous optimization, the high cost of search-based LON construction precludes the assembly of instance sets that systematically cover the LON feature space, leaving the LON feature--performance link largely unexamined.
To address this, we propose a search-free LON construction for \textit{Max-Set of Gaussians} (MSG) landscapes, a problem class with explicitly tunable multimodality.
Our alternative BoA definition bypasses search-based BoA identification, enabling low-cost LON construction.
We further leverage \textit{Novelty Search} (NS) to explore the parameter space of MSG landscapes, generating instances with diverse graph topologies.
Our experiments show that the alternative BoAs closely align with gradient-based BoAs, and that NS successfully generates MSG landscapes with varied search difficulty and connectivity patterns among optima.
Finally, on the instances generated by our methods, LON features accurately predict the success rates of two well-established evolutionary algorithms.
\end{abstract}

\section{Introduction} 
\let\oldthefootnote\thefootnote
\renewcommand{\thefootnote}{}
\footnotetext{$^{\dagger}$These authors contributed equally to this work.}
\let\thefootnote\oldthefootnote
Understanding the interplay between fitness landscapes and evolutionary algorithm performance is a long-standing challenge in evolutionary computation.
\textit{Local Optima Networks} (LONs)~\cite{LONs} provide a foundation for tackling this challenge by representing a search space's global structure as a coarse-grained graph, where nodes correspond to local optima and edges encode transitions between their \textit{Basins of Attraction} (BoAs). 
LONs and their variants enable intuitive visualization of high-dimensional search spaces and are powerful tools for estimating search difficulty and selecting appropriate algorithms~\cite{LON_Pagerank,PLOSnet_AAS,CPLOSnet,APLOSnet}.

While LONs have been primarily developed for discrete search spaces, recent studies have extended them to continuous search spaces~\cite{LON_Continuous,LON_Real,LON_BBOB,LON_scalable}.
However, LON construction in the continuous domain remains computationally demanding: identifying local optima and approximating BoA connectivity requires iterative execution of the \textit{Basin-Hopping} algorithm~\cite{BH} or the $(1+\lambda)$-\textit{Evolution Strategy} (ES)~\cite{mu_lambda}.
This search-based construction cost makes it impractical to assemble a comprehensive benchmark set that systematically covers the LON feature space.
Consequently, whether LON features quantitatively predict the performance of continuous optimizers has been examined only on individual benchmark problems and remains an open question.

In this study, we propose a framework\footnote{GitHub: \url{https://github.com/KPIMZT/Diverse\_LON}\\Zenodo: \url{https://doi.org/10.5281/zenodo.19630354}} that addresses these issues within the restricted setting of \textit{Max-Set of Gaussians} (MSG) landscapes---a parametric problem class defined as the pointwise maximum of a set of Gaussian functions, in which the local optima are a subset of the Gaussian centers.
Our framework consists of two components.
The first is a search-free LON construction for MSG landscapes: the region dominated by each Gaussian provides an alternative definition of the associated BoA, allowing LONs to be built without any iterative search.
The second is the exploration of the parameter space of MSG landscapes via \textit{Novelty Search} (NS)~\cite{NS}, which generates instances that maximize diversity in the LON feature space.
Using the dataset generated by NS, we investigate the LON feature--performance link empirically.
We note that our LON construction is specific to MSG landscapes and does not accelerate LON construction for arbitrary continuous problems.
Nevertheless, this framework provides a systematic dataset of feature vector--algorithm performance pairs.
We expect this dataset to serve as a foundation for landscape-aware optimization.

To assess our framework, we address the following research questions:
\begin{itemize}
    \item \textbf{RQ1}: To what extent do the BoAs of our definition coincide with those obtained by \textit{Gradient Descent} (GD)?
    \item \textbf{RQ2}: Can NS generate MSG landscapes whose LONs exhibit diverse graph topological features?
    \item \textbf{RQ3}: How do LON features extracted from MSG landscapes quantitatively predict the performance of continuous optimizers?
\end{itemize}

Our main contributions are as follows:
\begin{itemize}
    \item We introduce a fast LON construction for MSG landscapes, bypassing iterative execution of the search algorithm.
    \item We develop a novelty-based method for generating MSG landscapes that expands the reachable region in the space of LON features.
    \item We provide empirical evidence that LON features predict success rates of \textit{Covariance Matrix Adaptation} ES (CMA-ES)~\cite{CMA-ES} and \textit{Differential Evolution} (DE)~\cite{DE} with cross-validated ${R^2}$ ranging from 0.64 to 0.94.
\end{itemize}
\section{Preliminaries}
\subsection{Max-Set of Gaussians Landscape Generator}
The \textit{Max-Set of Gaussians} (MSG) landscape~\cite{MSG} is a parametric problem class constructed from multiple multivariate Gaussian functions.
In this paper, each instance is defined on the design variable space 
$X = [0,1]^d$ and is composed of a set of $m$ Gaussians 
$G = \{g_i\}_{i=1}^{m}$. 
Each Gaussian is defined as
\begin{equation}
    g_i(\sol{x})
    := w_i \exp \Big( -\tfrac{1}{2}
       (\sol{x}-\sol{c}_i)^\mathsf{T} \sol{\Sigma}_i^{-1} 
       (\sol{x}-\sol{c}_i) \Big),
    \label{eq:gaussian}
\end{equation}
where $w_i \in [0,1]$ is the peak height of $g_i$, 
$\sol{c}_i \in [0,1]^d$ is its center, and 
$\sol{\Sigma}_i \in \mathbb{R}^{d \times d}$ is its covariance matrix.
The objective value of $\sol{x}\in X$ is given by the pointwise maximum of 
the $m$ Gaussians:
\begin{equation}
    f(\sol{x}) := \max_{g_i \in G} g_i(\sol{x}).
    \label{eq:msg-objective}
\end{equation}
In this work, we treat $f$ as a function to be maximized.

\begin{figure}[b]
    \centering
    \begin{minipage}{0.8\textwidth}
        \centering
        \begin{subfigure}{0.45\linewidth}
            \centering
            \includegraphics[width=\linewidth]{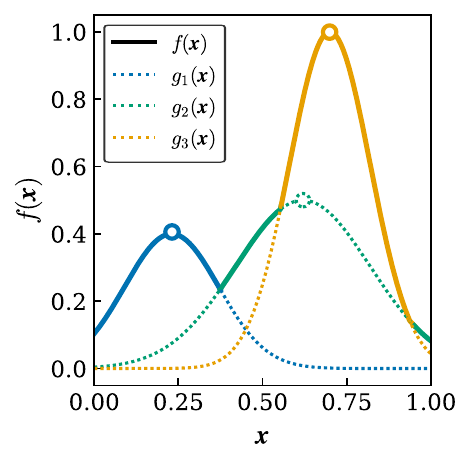}
            \caption{Surface.}
            \label{fig:1dsurface}
        \end{subfigure}%
        \hfill%
        \begin{subfigure}{0.45\linewidth}
            \centering
            \includegraphics[width=\linewidth]{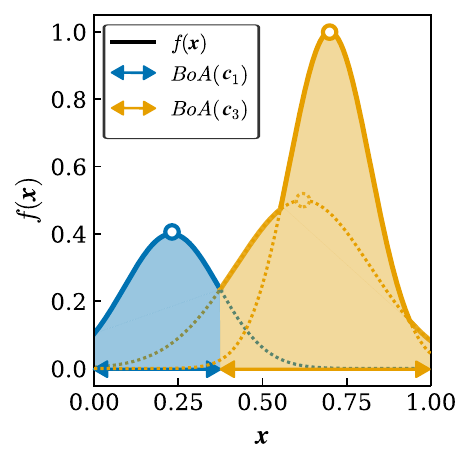}
            \caption{Basins of attraction.}
            \label{fig:1dBoA}
        \end{subfigure}%
    \end{minipage}
    \caption{An MSG landscape and its basins of attraction ($d=1$, $m=3$).}
    \label{fig:1d}
\end{figure}

An example is shown in~\cref{fig:1d}(\subref{fig:1dsurface}).
This instance is built from $m=3$ Gaussians $g_1, g_2, g_3$, shown as dashed lines, with the solid line representing the resulting objective function $f$.
The set of local optima $\LO$ consists of those Gaussian centers $C=\{\sol{c}_i\}_{i=1}^{m}$ that are not dominated by any other Gaussian at their own center, and is defined in~\cref{localoptima}:
\begin{equation}
    \LO :=
    \left\{ \sol{c}_i \in C \ \Big|\ \forall j \neq i:\ g_i(\sol{c}_i) > g_j(\sol{c}_i) \right\}.
    \label{localoptima}
\end{equation}

Because $\LO$ is by definition a subset of the Gaussian centers $C$, enumerating all local optima requires $m$ function evaluations. 
Even accounting for the fact that each function evaluation requires evaluating up to $m$ Gaussians, the overall number of Gaussian evaluations is at most $O(m^2)$ and is independent of the dimensionality $d$ of the design variable space.

\subsection{Local Optima Networks}
A \textit{Local Optima Network} (LON)~\cite{LONs} is a weighted directed graph $G_{\mathrm{LON}} = (\LO, E_w)$ whose nodes are the local optima and whose edges encode the connectivity between their BoAs.
The BoA of a local optimum $\sol{x}^*$ is the set of initial points from which a search procedure converges to $\sol{x}^*$:
\begin{equation}
    \mathrm{BoA}(\sol{x}^*) := 
    \{ \sol{x} \in X \mid \gamma(\sol{x}) = \sol{x}^* \},
\end{equation}
where $\gamma: X \to \LO$ denotes the local search operator.
Note that the BoA depends on the choice of search operator $\gamma$; different operators (e.g., hill climbing, GD, BFGS) may yield different BoAs for the same local optimum.

Several edge definitions have been proposed in the literature.
The most widely used is the \textit{escape edge}~\cite{escape_edge}.
Given a local optimum $\sol{x}_i^* \in \LO$, if a random perturbation followed by a local search leads to another local optimum $\sol{x}_j^* \in \LO$, a directed edge $(\sol{x}_i^*, \sol{x}_j^*)$ is added to $E_w$.
The edge weight is set to the estimated transition probability.

\subsection{Novelty Search}
Unlike conventional search algorithms, NS~\cite{NS} pursues solutions with diverse features.
NS operates on a \textit{genotype}--\textit{phenotype} mapping $\phi: X \to Y$, where a genotype $\sol{x} \in X$ is a design variable vector and its phenotype $\sol{y} = \phi(\sol{x}) \in Y$ is a feature vector characterizing the solution.
NS considers the novelty of a solution $\sol{z} = (\sol{x}, \sol{y})$ in the feature space $Y$ as the selection criterion, instead of an objective value. 
Apart from this criterion, the standard variation operators of evolutionary algorithms can be used to update the population.

In this work, our implementation of NS is based on $(\mu+\lambda)$-ES~\cite{mu_lambda}.
At each generation, the $(\mu+\lambda)$-ES produces $\lambda$ offspring from the $\mu$ parents, and then selects the next $\mu$ parents from the combined set of $\mu + \lambda$ solutions $Z_{\mu+\lambda} = Z_\mu \cup Z_\lambda$.
Each solution is assigned a novelty score computed relative to the reference set $Z' = Z_{\text{archive}} \cup Z_{\mu+\lambda}$, where $Z_{\text{archive}}$ is the novelty archive---a set of previously visited solutions with high novelty.
The novelty score is defined as 
\begin{equation}
\mathrm{novelty}(\sol{z}, Z') 
:= \frac{1}{k} \sum_{i=1}^{k} 
\mathrm{dist}(\sol{z}, \sol{z}^{\text{neighbor}}_{i}),
\label{NS}
\end{equation}
where $\sol{z}^{\text{neighbor}}_i \in Z'$ is the $i$-th nearest neighbor of $\sol{z}$ in the feature space, and $\mathrm{dist}(\cdot,\cdot)$ is a distance function on $Y$, typically the Euclidean or Mahalanobis distance.
Intuitively, solutions that exist far from others in the feature space obtain high novelty scores.
Any offspring solution $\sol{z}$ whose novelty score exceeds a predefined threshold $\tau$ is added to the novelty archive $Z_{\text{archive}}$ and is used when computing novelty scores in later generations.

\section{Proposed Methods}
\subsection{Search-Free LON Construction Based on Alternative BoAs}
To construct LONs on MSG landscapes without running search methods, we propose an alternative definition of BoAs.
First, we partition the design variable space into $m$ regions, one per Gaussian. The region associated with $g_i$ is defined as $R_i = \{\sol{x} \in [0,1]^d \mid f(\sol{x}) = g_i(\sol{x})\}$, i.e., the set of points at which $g_i$ attains the maximum.\footnote{Boundary points where multiple Gaussians attain the maximum form a measure-zero set and are assigned arbitrarily.}
Next, as illustrated in~\cref{fig:1d}(\subref{fig:1dBoA}), whenever the center $\sol{c}_i$ of $g_i$ is dominated by another Gaussian $g_j$ (i.e., $c_i \in R_j$), we merge $R_i$ into $R_j$.
Because each merge strictly increases the $f$-value of the representative center ($f(\sol{c}_j) > f(\sol{c}_i)$), this process cannot cycle and terminates at a local optimum.
Let $\sol{x}^\ast = \sol{c}_i \in \LO$ be a local optimum and $I(\sol{x}^\ast)\subseteq\{1, \ldots,m\}$ denote the set of indices of the Gaussians merged into the region corresponding to $\sol{x}^\ast$. 
In our definition the BoA of $\sol{x}^\ast$ is $\mathrm{BoA}_{ours}(\sol{x}^\ast) = \bigcup_{i \in I(\sol{x}^\ast)} R_i$.

Edge weights follow the standard escape-edge construction. 
For each local optimum $\sol{x}^\ast$, we sample $s$ candidates uniformly from the ball of radius $r$ centered at $\sol{x}^\ast$ (clipped to $[0,1]^d$), assign each sample to a BoA via the procedure above, and set the weight of each outgoing edge to the fraction of samples landing in the corresponding target BoA. Under this definition, enumerating all local optima requires $m$ function evaluations, and approximating inter-optimum transitions requires $m \cdot s$ function evaluations.

\subsection{MSG Landscape Generation via Novelty Search}
We apply novelty search using the MSG generator parameters $\sol{\theta}$ as the genotype and LON features as the phenotype.
The full parameter vector $\sol{\theta}$ consists of $(w_i, \sol{c}_i, \bm{\Sigma}_i)$ for $i = 1, \dots, m$, giving a dimension of $m\left(1 + d + \tfrac{d(d+1)}{2}\right)$.
To keep this search space tractable, we follow previous work~\cite{MSG_para} and reduce the parameter count in two ways: we fix the Gaussian centers $C$ by sampling them from a Sobol' sequence~\cite{sobol1967}, and we restrict each Gaussian to be isotropic (i.e., $\bm{\Sigma}_i = \sigma_i^2 \sol{I}_d$).
The remaining search variables are the peak height and scale of each Gaussian, giving a $2m$-dimensional genotype $\sol{\theta} = (w_1,\dots, w_m, \sigma_1,\dots,\sigma_m)$.
While these reductions restrict representable landscapes, we have previously shown~\cite{MSG_para} that MSG landscapes under the same restrictions can emulate most BBOB function classes~\cite{BBOB}; only landscapes with extreme ruggedness or strong anisotropy remain out of scope.

We define the phenotype mapping $\phi$ as follows: given a genotype $\sol{\theta}$, we build the MSG landscape $f_{\sol{\theta}}$, construct its LON, and take the graph features of the LON as the phenotype $\sol{y} = \phi(\sol{\theta})$.
The pseudocode of our framework is shown in \cref{alg:NSMSG}.

\begin{algorithm}[th!]
\SetArgSty{textnormal}
\fontsize{8pt}{8pt}\selectfont
\LinesNotNumbered
\DontPrintSemicolon
\caption{MSG landscape generation framework using $(\mu+\lambda)$-NS}
\label{alg:NSMSG}
\SetKwProg{Fn}{Procedure}{:}{}
\SetKwFunction{main}{($\mu$+$\lambda$)-NS}
\SetKwFunction{init}{initialize}
\SetKwFunction{evol}{reproduce}
\Fn{\main{$d,\mu, \lambda, t_{\max}, \tau, k, \alpha_w, \alpha_\sigma,\sigma_{\min}, \sigma_{\max}, \phi, m$}}
{
    $Z_\mu \gets \init(d,\mu, \sigma_{\min}, \sigma_{\max}, \phi, m)$\;
    $Z_{\text{all}},\; Z_{\text{archive}} \gets Z_\mu$\;
    \For{$t \gets 1$ \KwTo $t_{\max}$}
    {
        $Z_\lambda \gets \evol(\lambda, Z_\mu, \alpha_w, \alpha_\sigma)$\;
        $Z_{\mu+\lambda} \gets Z_\mu \cup Z_\lambda$;\quad $Z' \gets Z_{\text{archive}} \cup Z_{\mu+\lambda}$\;
        $Z_{\text{archive}} \gets Z_{\text{archive}} \cup \{\sol{z} \in Z_\lambda \mid \mathrm{novelty}(\sol{z}, Z') > \tau\}$\;
        $Z_{\text{all}} \gets Z_{\text{all}} \cup Z_\lambda$\;
        $Z_\mu \gets$ top-$\mu$ of $Z_{\mu+\lambda}$ by $\mathrm{novelty}(\cdot, Z')$\;
    }
    \Return $Z_{\text{all}}$\;
}
\;
\Fn{\init{$d,\mu, \sigma_{\min}, \sigma_{\max}, \phi, m$}}
{
    $C \gets m$ points sampled from $[0,1]^d$ via a Sobol' sequence\;
    \For{$i \gets 1$ \KwTo $\mu$}
    {
        $w_j \sim \mathcal{U}[0,1],\ \sigma_j \sim \mathcal{U}[\sigma_{\min}, \sigma_{\max}]$ for $j = 1, \dots, m$\;
        $\sol{\theta}_i \gets (w_1,\dots, w_m, \sigma_1,\dots,\sigma_m)$\;
    }
    \Return $\{(\sol{\theta}_i, \phi(\sol{\theta}_i))\}_{i=1}^{\mu}$\;
}
\;
\Fn{\evol{$\lambda, Z_\mu, \alpha_w, \alpha_\sigma$}}
{
    \For{$i \gets 1$ \KwTo $\lambda$}
    {
        $(\sol{\theta}_p, \cdot) \gets$ uniformly random sample from $Z_\mu$\;
        $\sol{\theta}'_i \gets \sol{\theta}_p + (\alpha_w \sol{\epsilon}_w,\, \alpha_\sigma \sol{\epsilon}_\sigma),\ \sol{\epsilon}_w, \sol{\epsilon}_\sigma \sim \mathcal{N}(\sol{0}, \sol{I})$\;
    }
    \Return $\{(\sol{\theta}'_i, \phi(\sol{\theta}'_i))\}_{i=1}^{\lambda}$\;
}
\end{algorithm}

\section{Experiments}
\subsection{Common Settings}
This subsection describes the common settings shared across all experiments (RQ1--RQ3), specifically the parameters of MSG landscapes and LONs.

\subsubsection{Constructing MSG Landscapes:}
We set the design variable space of MSG landscapes to $[0,1]^d$ with dimensionality $d \in \{2, 5, 10\}$.
Each instance is composed of $m = 50 \times d$ Gaussians.
The center $\sol{c}$ of each Gaussian is sampled on $[0,1]^d$ using a Sobol' sequence and fixed.
The peak height $w$ is bounded to $[0,1]$, and scale $\sigma$ is bounded to $[\sigma_{\min}=r/4,\sigma_{\max}=3r]$ where $r = \sqrt[d]{1/m}$ is the side length obtained by partitioning the design variable space into $m$ equal hypercubes.

\subsubsection{Constructing LONs:}
When constructing LONs, we reuse $r=\sqrt[d]{1/m}$ as the sampling radius and sample $s = 500 \times d$ points per local optimum to estimate the neighboring BoAs.
We use the metrics in \cref{tab:metrics} as LON features. 
Most of these features are typically computed from a \textit{monotonic} LON~\cite{LON_mono}, i.e., a LON containing only improving edges. 
However, for \texttt{funnel\_size\_opt}, we further prune the network to keep only the highest-weight outgoing edge per node. 

\begin{table}[t!]
    \centering
    \caption{LON feature metrics}
    \label{tab:metrics}
    \footnotesize
    \begin{tabularx}{\linewidth}{lX}
        \toprule
        Metric & Description \\
        \midrule
        \texttt{num\_nodes}           & Number of nodes. \\
        \texttt{edge\_density}        & Ratio of the number of edges to that of a complete graph. \\
        \texttt{num\_sinks}           & Number of nodes with out-degree 0 (sink nodes), corresponding to the number of funnels. \\
        \texttt{avg\_path\_opt}       & Average shortest-path length from each node to the node corresponding to the global optimum (optimal node). \\
        \texttt{avg\_path\_sinks}     & Average shortest-path length from each node to its nearest sink. \\
        \texttt{in\_strength\_opt}    & Weighted in-degree of the optimal node. \\
        \texttt{in\_strength\_sinks}  & Average weighted in-degree across sink nodes. \\
        \texttt{funnel\_size\_opt}    & Fraction of nodes contained in the funnel whose sink is the optimal node (optimal funnel). \\
        \bottomrule
    \end{tabularx}
\end{table}

\subsection{Validation of the Alternative LONs via BoA Comparison}
\subsubsection{Experimental Setup:}

RQ1 asks whether our alternative BoA definition yields LONs that reflect the landscape structure.
Since the local optima of MSG landscapes are known analytically (\cref{localoptima}), the only component requiring validation is the BoA assignment. 
We therefore evaluate our LONs by measuring how closely $\text{BoA}_{\textit{ours}}$ coincides with $\text{BoA}_{\textit{GD}}$, i.e., BoAs of \textit{Gradient Descent} (GD), a standard geometric reference.
On an MSG landscape, the gradient at $\sol{x}$ is determined by the Gaussian $g_k$ that attains the maximum at $\sol{x}$.
A standard GD update may overshoot the basin boundary when $\sigma_k$ is small.
To prevent this, we cap the step size so that it never moves past the dominating Gaussian's center:
\begin{equation}
    \sol{x} \leftarrow \sol{x} + \min\!\left(\frac{\eta\, g_k(\sol{x})}{\sigma_k^2},\; 1\right) \cdot \bigl(\sol{c}_k - \sol{x}\bigr),
    \label{gd_update}
\end{equation}
where $\sigma_k$ and $\sol{c}_k$ are scale and center of $g_k$.

We generate $5{,}000 \times d$ starting points in $[0,1]^d$ via a Sobol' sequence and run GD from each one, with learning rate $\eta = 0.01$ and a maximum of $T = 2{,}000$ steps.
Points leaving $[0,1]^d$ during an update are clipped back to the boundary.
We estimate $\text{BoA}_{\textit{GD}}$ from the convergence point.
We use \texttt{difference\_rate} as the evaluation metric, defined as the fraction of starting points assigned to different basins by $\text{BoA}_{\textit{GD}}$ and $\text{BoA}_{\textit{ours}}$.
For each dimensionality $d$, we randomly generate $100$ MSG landscapes as the comparison set.

\begin{figure*}[t]
    \centering
    \includegraphics[width=\textwidth]{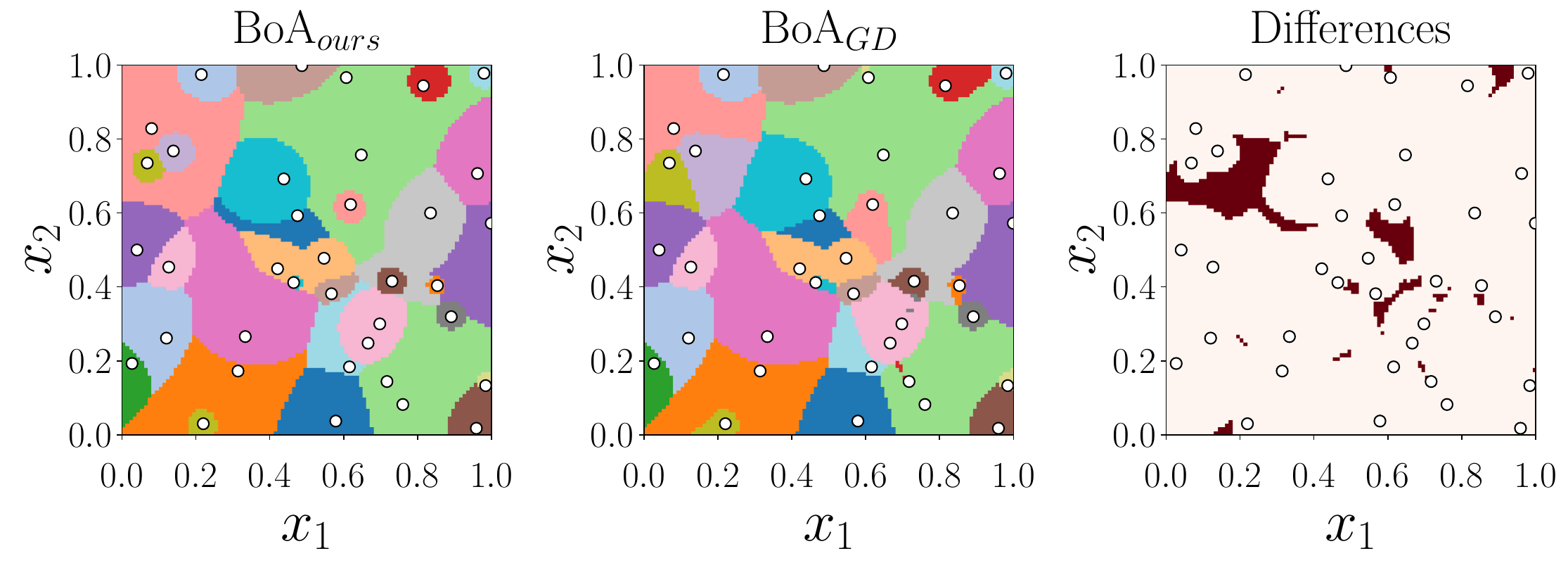}
    \caption{Comparison of $\text{BoA}_{\textit{ours}}$ and
    $\text{BoA}_{\text{GD}}$ for the instance with the highest
    \texttt{difference\_rate} ($6.49\%$) at $d=2$.}
    \label{fig:BoAs}
\end{figure*}

\subsubsection{Results and Interpretation:}
The experimental results show that the median \texttt{difference\_rate} is $2.04\%$ for $d=2$, $0.46\%$ for $d=5$, and $0.07\%$ for $d=10$.
\cref{fig:BoAs} compares $\text{BoA}_{\textit{ours}}$ and $\text{BoA}_{\textit{GD}}$ for the instance that yielded the highest \texttt{difference\_rate} ($6.49\%$) in the experiment with $d=2$.
The differences between the two BoAs are concentrated near the boundary of a basin that is nested inside another basin.
In such a region, GD assigns points to the inner (nested) basin, whereas our definition assigns them to the outer (enclosing) basin.
In other words, $\text{BoA}_{\textit{ours}}$ matches $\text{BoA}_{\textit{GD}}$ over most of the space and only deviates where one basin is nested inside another.
As the dimensionality $d$ increases, the relative volume of these boundary regions shrinks, so uniformly sampled points are less likely to fall near a boundary.
This may explain the decrease in \texttt{difference\_rate} in higher dimensions.

\subsubsection{Answer to RQ1:}
The proposed BoA definition closely approximates the GD-based BoA, with median \texttt{difference\_rate} decreasing from $2.04\%$ at $d=2$ to $0.07\%$ at $d=10$.
Differences are concentrated near the boundaries of nested basins, where one basin is contained within another.
These results support the validity of our BoA definition for LON construction on MSG landscapes.

\subsection{Comparative Evaluation of Diversity}
\subsubsection{Experimental Setup:}
\begin{figure}[t!]
\centering
\begin{subfigure}{0.32\linewidth}
    \includegraphics[width=\linewidth]{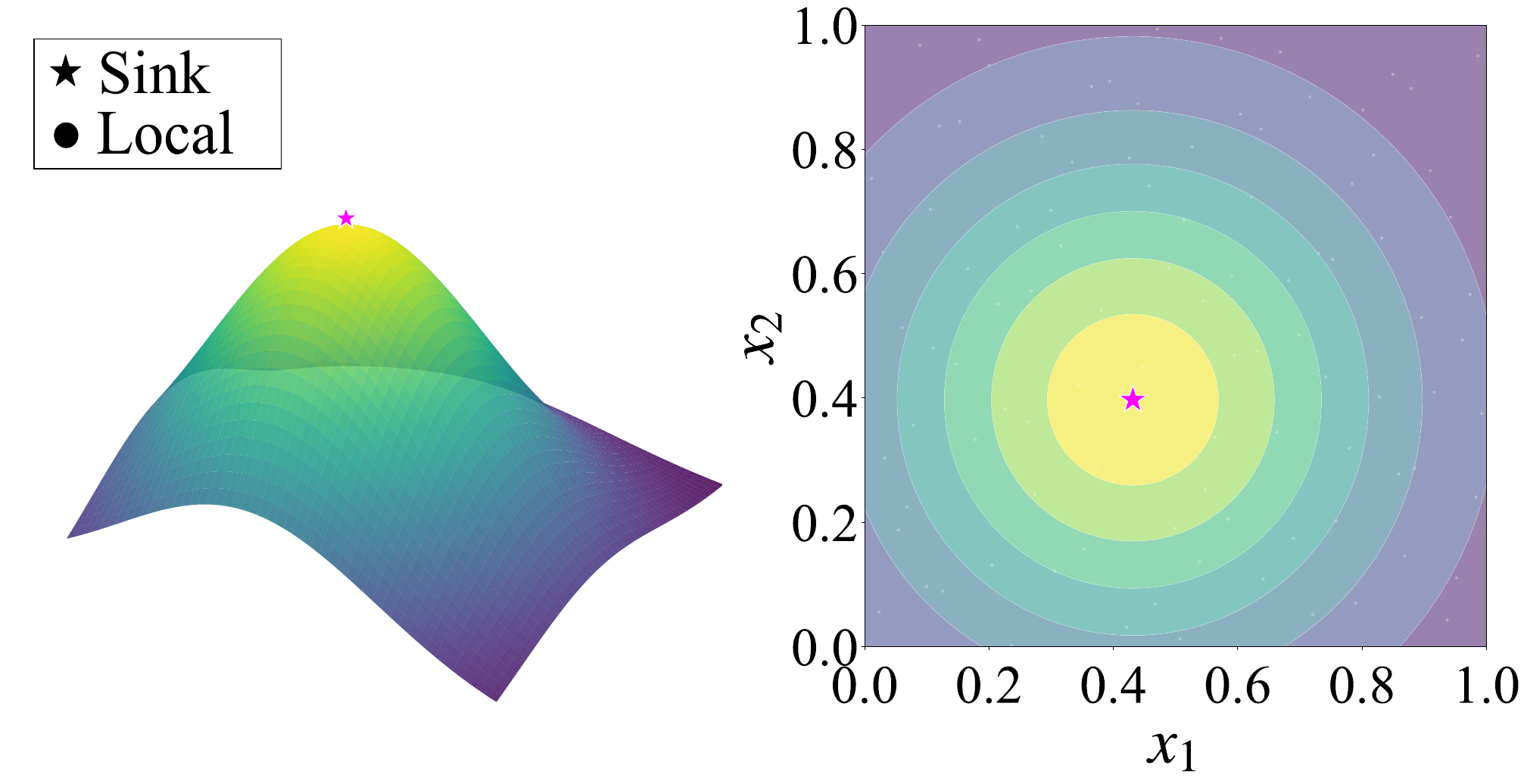}
    \caption{Unimodal instance}
\end{subfigure}
\hfill
\begin{subfigure}{0.32\linewidth}
    \includegraphics[width=\linewidth]{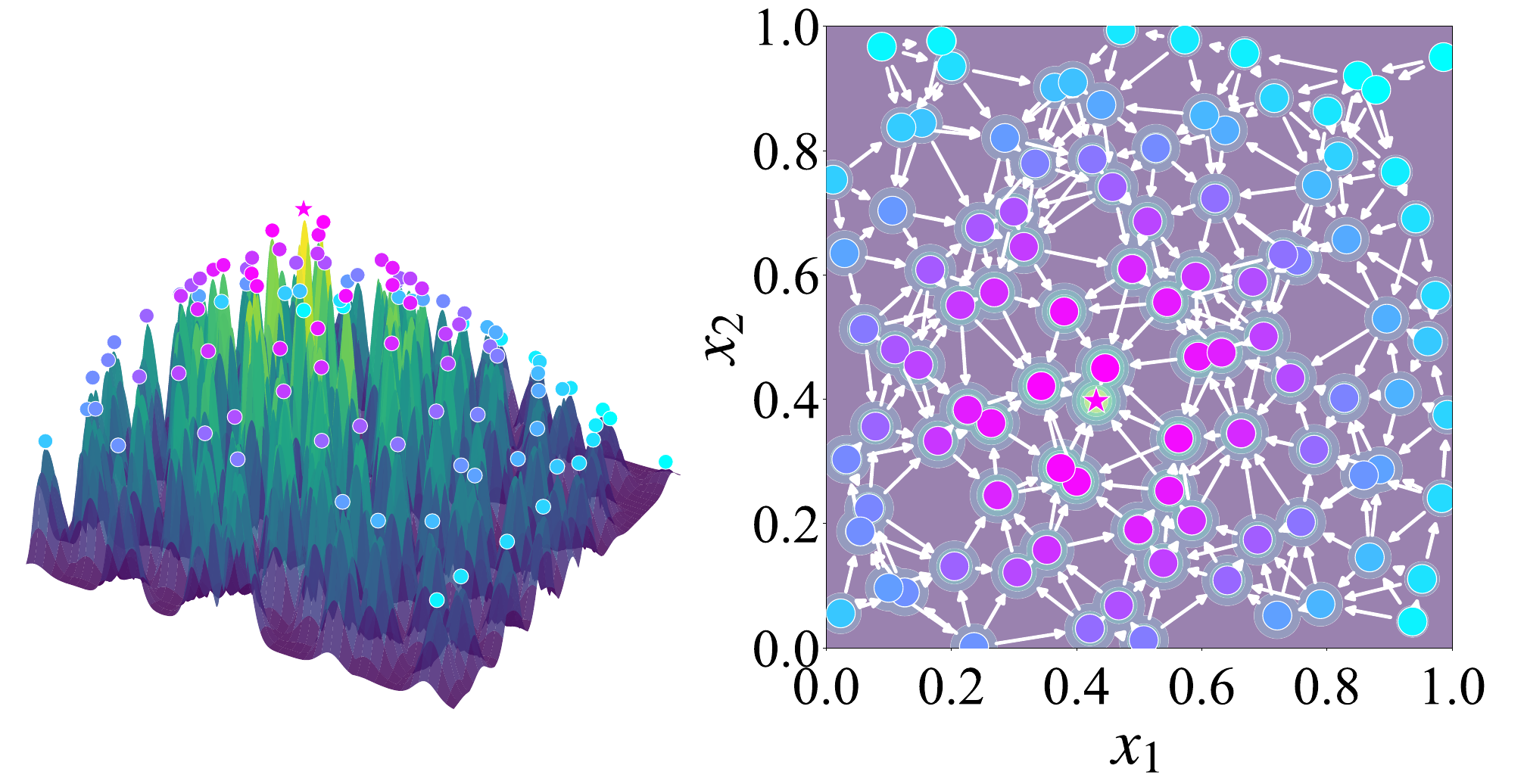}
    \caption{Uni-sink instance}
\end{subfigure}
\hfill
\begin{subfigure}{0.32\linewidth}
    \includegraphics[width=\linewidth]{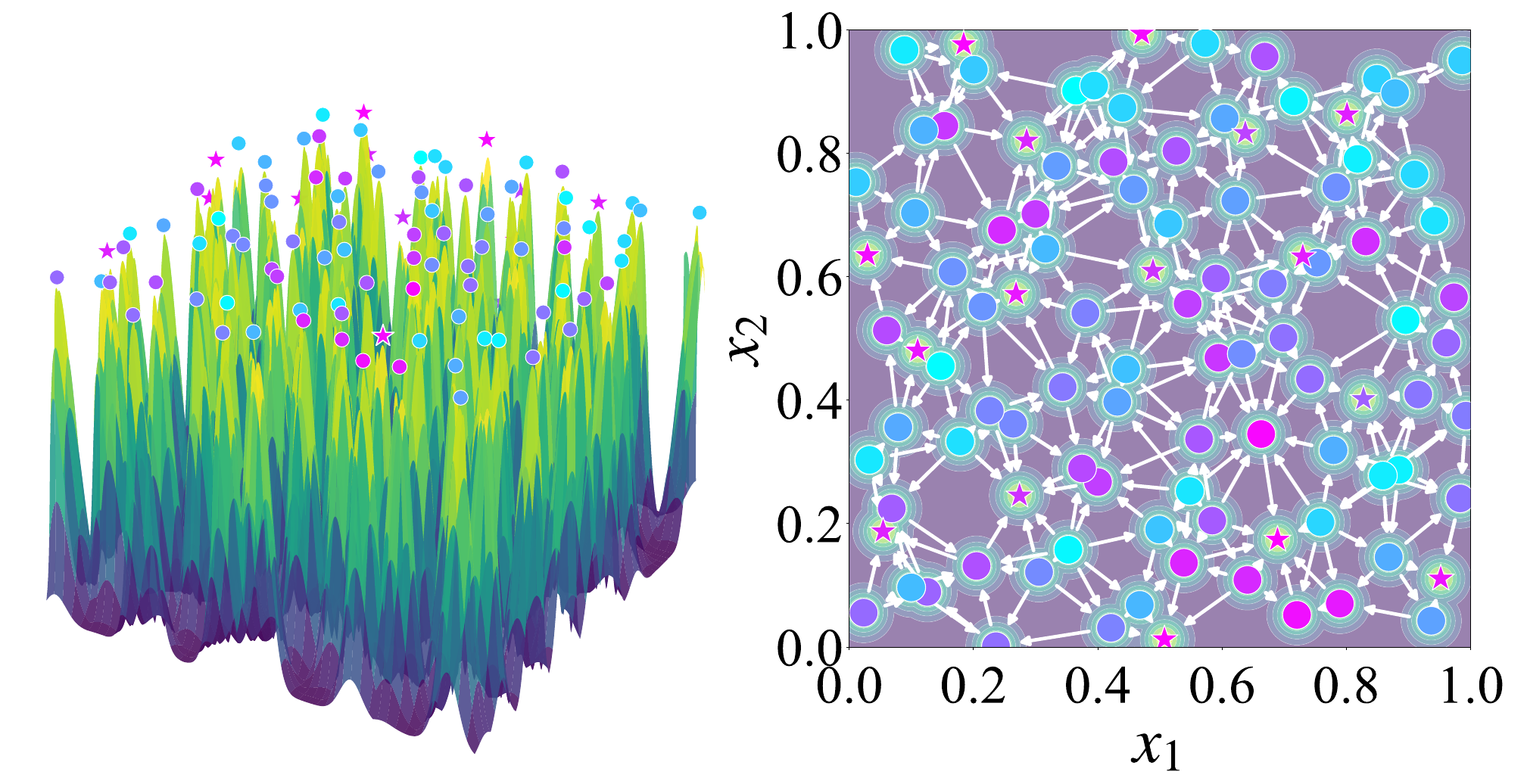}
    \caption{Multi-sink instance}
\end{subfigure}
\caption{MSG landscapes included in the initial population of $\texttt{NS}^{+}$ and their corresponding LONs.}
\label{archetype2d}
\end{figure}

RQ2 evaluates the effectiveness of our MSG landscape generation.
We compare instances generated by NS (\texttt{NS}) and by random search (\texttt{Random}) in the LON feature space.
We also assess the effect of the initial population on \texttt{NS}.
The standard \texttt{NS} starts from a random initial population.
$\texttt{NS}^{+}$ replaces $3$ solutions with hand-designed, structurally distinct MSG landscapes shown in \cref{tab:archetype} and \cref{archetype2d}.

For both \texttt{NS} and $\texttt{NS}^{+}$, we use parent size $\mu = 20$, offspring size $\lambda = 100$, and $t_{\max} = 100$ generations, yielding $\mu + t_{\max} \times \lambda = 10{,}020$ MSG landscapes in total.
The mutation step sizes are set to $\alpha_w = 0.1$ and $\alpha_\sigma = 0.05$.
The novelty score is calculated as the average Euclidean distance to the $k = 15$ nearest neighbors in the feature space.
The novelty threshold is set to $\tau = 0.05$ and is dynamically adjusted based on the number of archive additions per generation, following the approach of existing studies~\cite{NS2008,NS_em}. 
In this work, the threshold is evaluated every $4$ generations: if the total number of archive additions over 4 generations exceeds 30, the threshold is multiplied by 1.05; if the number of additions is 0, it is multiplied by 0.95.

The feature space is the 2-dimensional space spanned by \texttt{num\_nodes} and \texttt{funnel\_size\_opt}.
\texttt{num\_nodes} is normalized by the number of Gaussians $m$.
The \texttt{Random} set contains the same number of MSG landscapes, each generated by independent uniform sampling of the parameters.
We measure the coverage in the feature space to compare \texttt{NS} and \texttt{Random}.
For calculating coverage, the feature space is partitioned into $30 \times 30 = 900$ cells. 
The coverage is computed as the fraction of cells containing at least one instance.
For each dimensionality $d$, we run 10 trials with different seeds and evaluate the final cumulative coverage over all generated instances across \texttt{Random}, \texttt{NS}, and $\texttt{NS}^{+}$.

\begin{table}[t!]
    \centering
    \caption{MSG landscapes included in the initial population of $\texttt{NS}^{+}$.}
    \label{tab:archetype}
    \footnotesize
    \begin{tabularx}{\linewidth}{lX}
        \toprule
        Archetype & Parameter settings of the heights $w$ and scales $\sigma$ \\
        \midrule
        Unimodal   & One Gaussian $g_i$ has $w_i = 1, \sigma_i = \sigma_{\max}$; other Gaussians $g_{j\neq i}$ have $w_j = g_i(\sol{c}_j)-0.001, \sigma_j = \sigma_{\min}$. \\
        \addlinespace
        Uni-sink    & One Gaussian $g_i$ has $w_i = 1$; other Gaussians $g_{j\neq i}$ have $w_{j}$ that decrease linearly with distance from $\sol{c}_i$. All Gaussians have $\sigma = \sigma_{\min}$.\\
        \addlinespace
        Multi-sink  &All Gaussians start at $w_i = 1 + 0.01 \times \epsilon_i$ with $\epsilon_i \sim \mathcal{N}(0, 1)$, and are then normalized by $w_i \leftarrow w_i / \max_j w_j$. All Gaussians have $\sigma = \sigma_{\min}$. \\
        \bottomrule
    \end{tabularx}
\end{table}

\subsubsection{Results and Interpretation:}
\begin{figure}[t!] 
    \centering
    \includegraphics[width=\textwidth]{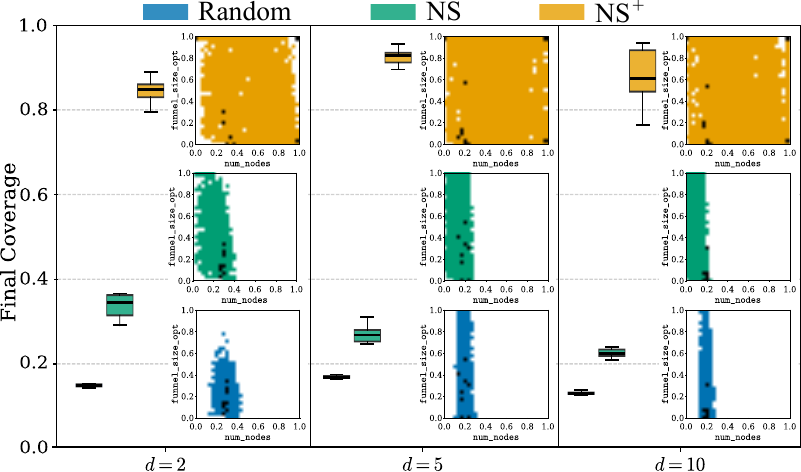}
    \caption{Box plots showing the final cumulative coverage of each instance group for $d \in \{2, 5, 10\}$. $10$ trials were conducted for each dimensionality. The heatmaps on the right show feature-space coverage from a single representative trial. Black cells contain the initial population; colored cells contain instances generated during the search (blue: \texttt{Random}, green: \texttt{NS}, orange: $\texttt{NS}^+$).}
    \label{allcoverage}
\end{figure}

\cref{allcoverage} shows the coverage of \texttt{Random}, \texttt{NS}, and $\texttt{NS}^{+}$ per dimensionality.
The box plots summarize coverage over 10 trials. 
The feature-space coverage shown on the right is taken from a single trial, where each colored cell indicates the presence of an MSG landscape.
In the box plots, \texttt{Random}, \texttt{NS}, and $\texttt{NS}^{+}$ are arranged from left to right.

\texttt{NS} covers a broader region of the feature space than \texttt{Random} at all dimensionalities.
This indicates that uniform sampling in parameter space does not imply uniformity in the LON feature space: 
the parameter-to-feature mapping is nonlinear, and MSG landscapes from \texttt{Random} concentrate in a limited region of the feature space.
\texttt{NS} directly maximizes feature-space diversity as the novelty score, compensating for this nonlinearity.

Adding the archetypes shown in \cref{archetype2d} to the initial population substantially improves the coverage, with median coverage exceeding $0.8$ at all dimensionalities.
A random initial population concentrates in a limited region, confining subsequent NS exploration to its vicinity.
In contrast, the initial population of $\texttt{NS}^{+}$ provides multiple well-separated anchor points in the feature space, 
and NS expands coverage by exploring their neighborhoods and the gaps between them.

\subsubsection{Answer to RQ2:}
NS generates MSG landscapes that reach regions of the LON feature space inaccessible to random generation, 
and initializing with archetypes further expands the reachable region to a median coverage above $0.8$.
However, NS alone achieves only moderate absolute coverage (median $0.2$–$0.4$ across dimensionalities), 
indicating that the choice of initial population is a critical factor, complementing novelty-driven variation.

\subsection{Correlation and Regression Analysis}
\subsubsection{Experimental Setup:}
RQ3 investigates whether LON features computed from our alternative BoA definition can predict the performance of optimization algorithms.
We begin by analyzing the relationship between LON features and algorithm performance using Spearman’s rank correlation. 
We then assess prediction accuracy with random forest regression using the \texttt{scikit-learn}~\cite{scikit-learn} implementation.
For the regression, we use a random forest with 200 trees, evaluating prediction accuracy via 10-fold cross-validation.
As test problems, we consider all $10{,}020$ MSG landscapes generated from a single $\texttt{NS}^{+}$ trial in RQ2.
As test algorithms, we use CMA-ES~\cite{CMA-ES} and DE~\cite{DE}, with the implementations provided by \texttt{pymoo}~\cite{pymoo}.

For CMA-ES, the initial mean is drawn from $\mathcal{U}[0,1]^d$. 
The parameter settings follow the default configuration in \texttt{pymoo}, with restarts disabled.
For DE, the initial population is generated in the same way.
We adopt DE/rand/1/bin with population size $10 \times d$, crossover rate $CR = 0.9$, and scaling factor $F=0.5$. Vector-level dither is applied, randomly perturbing $F$ for each individual.
These parameter values are commonly recommended settings from prior studies~\cite{DEconfig1,DEconfig2}.
Each algorithm is benchmarked across 31 independent trials, where a trial is considered successful if the coordinate error from the global optimum is below $10^{-2}$.
Trials terminate when the number of function evaluations, including the initial evaluation, reaches $1{,}000 \times d$.

We measure performance by two metrics: the success rate over all trials (\texttt{success\_rate}) and the mean number of evaluations until convergence or budget exhaustion (\texttt{conv\_time}). Convergence is declared when $\Delta x < 10^{-11}$ and $\Delta f < 10^{-11}$. If a trial terminates without converging, we record the number of evaluations at termination (i.e., $1{,}000 \times d$) as the \texttt{conv\_time}.

\begin{figure}[t!]
\centering
\begin{subfigure}{1.0\textwidth}
    \includegraphics[width=\linewidth]{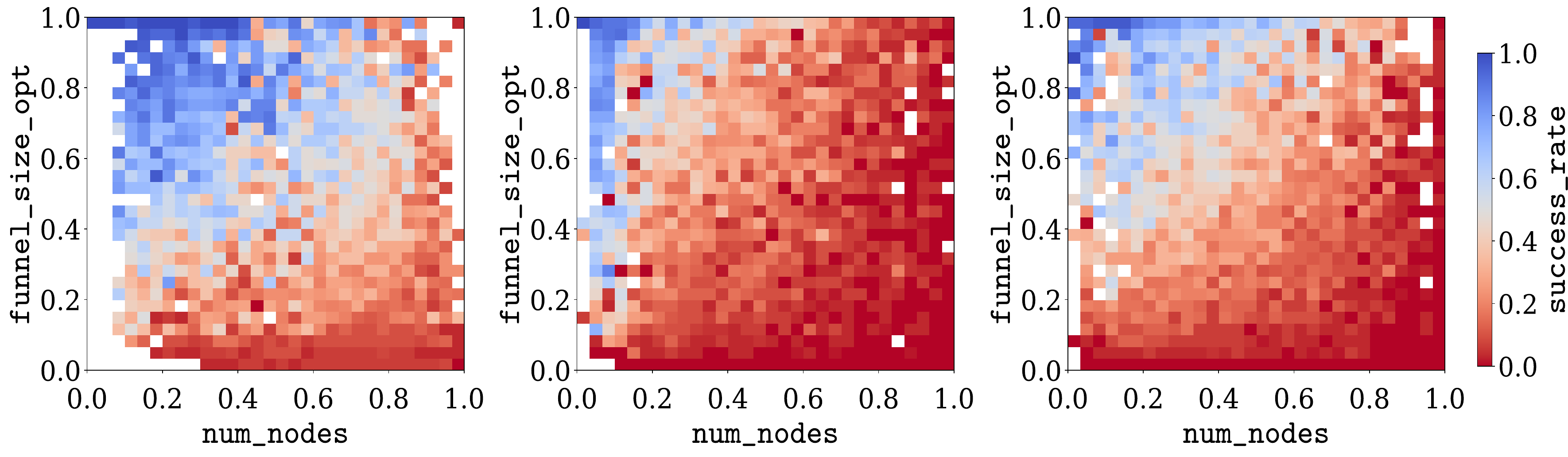}
    \caption{$d=2,5,10$, CMA-ES}
\end{subfigure}

\begin{subfigure}{1.0\textwidth}
    \includegraphics[width=\linewidth]{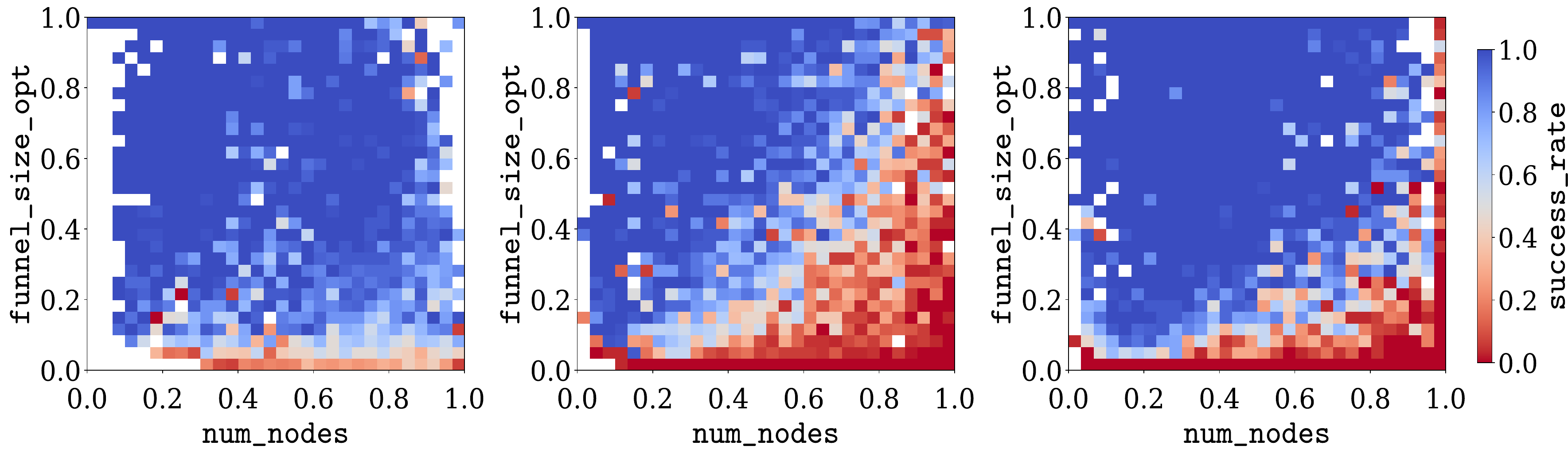}
    \caption{$d=2,5,10$, DE}
\end{subfigure}
\caption{\texttt{success\_rate} of CMA-ES and DE in each cell for $d \in \{2, 5, 10\}$. If there are multiple instances, the median value is displayed.}
\label{heatmap}
\end{figure}

\subsubsection{Results and Interpretation:}
We first discuss algorithm performance.
\cref{heatmap} shows the \texttt{success\_rate} of CMA-ES and DE for each cell in the feature space.
The generated instances cover a wide range of success rates, and the success rate tends to decrease toward the lower-right region for both algorithms.
This indicates that finding a global optimum becomes harder when instances have many local optima and a small optimal funnel.
This is consistent with the common understanding that landscapes with multiple competing funnels are difficult for evolutionary algorithms~\cite{LON_mono}.
Notably, the \texttt{success\_rate} distribution is more bimodal for DE than for CMA-ES: DE shows a sharp separation between regions of near-zero and near-one success rate with few intermediate values, whereas CMA-ES exhibits a more gradual transition.
This bimodality of DE further intensifies with dimensionality.
\begin{figure}[t!]
    \centering
    \includegraphics[width=\textwidth]{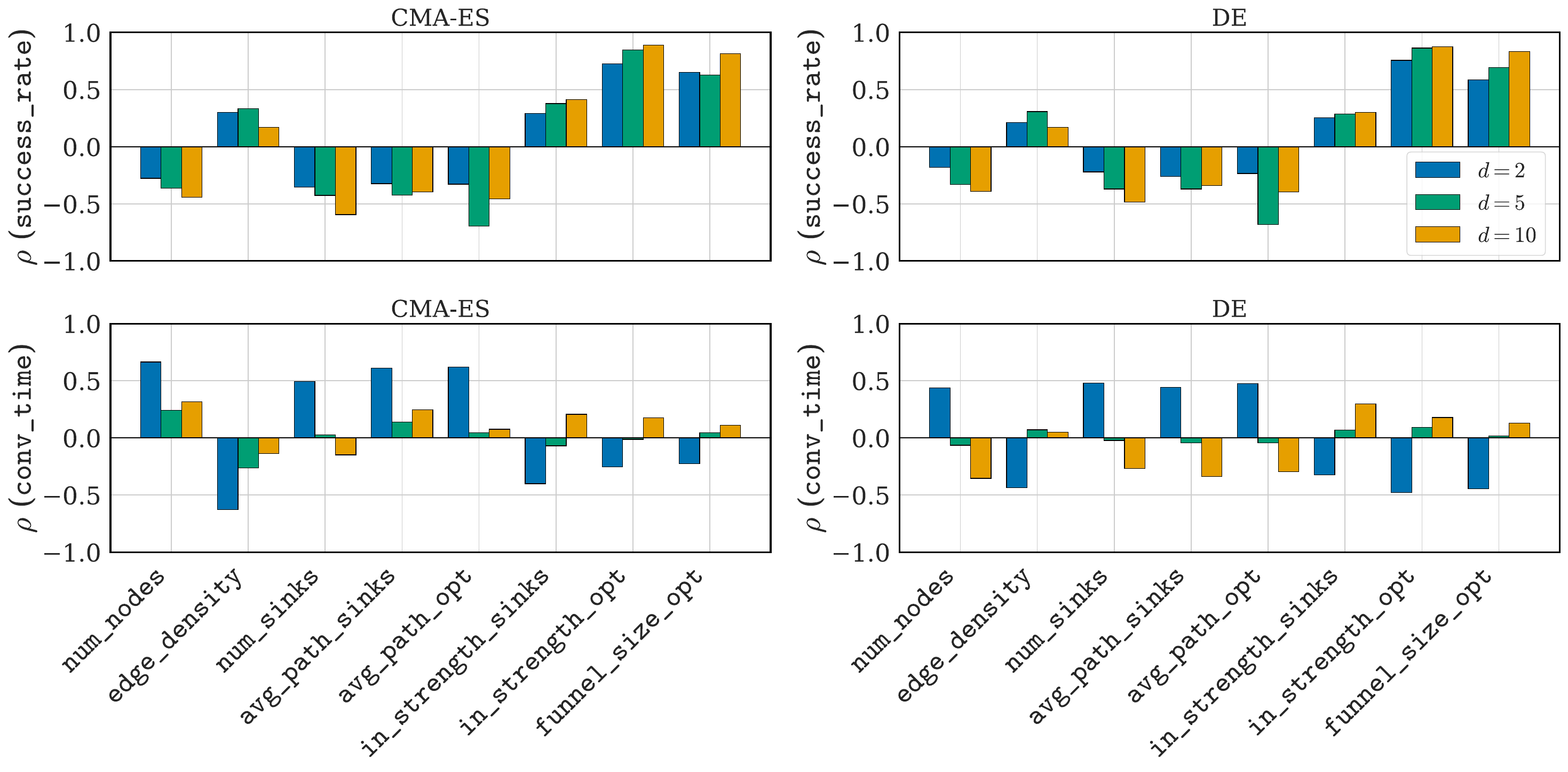}
    \caption{Spearman's rank correlation coefficient $\rho$ between LON features and the performance of CMA-ES and DE for $d \in \{2, 5, 10\}$.}
    \label{s_cor}
\end{figure}

Next, \cref{s_cor} shows Spearman's rank correlations between LON features and algorithm performance.
For \texttt{success\_rate}, CMA-ES and DE exhibit similar correlation patterns.
The correlation with \texttt{in\_strength\_opt} is high across all dimensionalities, with the magnitude increasing with $d$.
A similarly strong correlation is observed for \texttt{funnel\_size\_opt}, consistent with the interpretation that a larger optimal funnel makes the global optimum more reachable.
These two features correlate with \texttt{success\_rate} regardless of algorithm or dimensionality.
For \texttt{conv\_time}, many features correlate with performance at $d=2$ for both algorithms: in particular,  \texttt{num\_nodes} and \texttt{avg\_path\_$\ast$} indicate more local optima and longer network paths, which lead to longer convergence times.
However, few features show clear correlations at $d=5$ and $d=10$, suggesting that the LON features used here may not adequately capture the landscape structures governing convergence speed.

\cref{tab:rf_reg} shows the results of random forest regression for predicting algorithm performance from LON features.
For \texttt{success\_rate}, the cross-validated $R^2$ ($R^2_{\text{cv}}$) of CMA-ES is $0.775$, $0.851$, and $0.939$ at $d=2, 5, 10$, respectively.
For DE, the corresponding values are $0.638$, $0.838$, and $0.901$.
Prediction accuracy improves with dimensionality for both algorithms.
At first glance, the rising $R^2_{\text{cv}}$ with dimensionality appears counterintuitive.
As seen in \cref{heatmap}, the \texttt{success\_rate} distribution becomes increasingly bimodal as $d$ increases, especially for DE, where near-zero and near-one values dominate.
Under this bimodality, the regression task reduces to distinguishing easy from hard instances, so part of the rise in $R^2_{\text{cv}}$ at higher $d$ reflects a sharper separation of the target variable rather than stronger intrinsic predictability of LON features.
Nevertheless, the consistently lower $R^2_{\text{cv}}$ for DE compared to CMA-ES, despite DE's stronger bimodality, indicates that LON features capture aspects of CMA-ES behavior that go beyond the easy/hard distinction.

\begin{table}[t!]
    \centering
    \caption{Coefficient of determination $R^{2}_{\mathrm{cv}}$ for random forest regression. The numbers in parentheses indicate the standard deviation.}
    \label{tab:rf_reg}
    \footnotesize
    \begin{tabular*}{\linewidth}{@{\extracolsep{\fill}} c cc cc @{}}
        \toprule
        & \multicolumn{2}{c}{CMA-ES} & \multicolumn{2}{c}{DE} \\
        \cmidrule(lr){2-3} \cmidrule(lr){4-5}
        $d$ & \texttt{success\_rate} & \texttt{conv\_time} & \texttt{success\_rate} & \texttt{conv\_time} \\
        \midrule
        2  & 0.775(0.017) & 0.650(0.022) & 0.638(0.018) & 0.507(0.015) \\
        5  & 0.851(0.010) & 0.173(0.022) & 0.838(0.010) & 0.231(0.135) \\
        10 & 0.939(0.008) & 0.459(0.024) & 0.901(0.007) & 0.153(0.259) \\
        \bottomrule
    \end{tabular*}
\end{table}

For \texttt{conv\_time}, $R^2_{\text{cv}}$ is substantially lower: CMA-ES achieves $0.650$, $0.173$, and $0.459$, and DE achieves $0.507$, $0.231$, and $0.153$ at $d=2, 5, 10$, respectively.
These values show no consistent trend with dimensionality and are uniformly lower than those for \texttt{success\_rate}, suggesting that predicting convergence time is difficult with the present features.
This matches the correlation analysis above, which showed that the association between \texttt{conv\_time} and LON features weakens in higher dimensions.

\subsubsection{Answer to RQ3:}
LON features derived from our BoA definition enable random forest regression to predict the \texttt{success\_rate} of CMA-ES and DE with $R^2_{\text{cv}}$ ranging from $0.64$ to $0.94$, with \texttt{in\_strength\_opt} and \texttt{funnel\_size\_opt} showing the strongest correlations with \texttt{success\_rate}.
While part of the accuracy gain at higher $d$ is attributable to the increasing bimodality of \texttt{success\_rate}, the comparison between CMA-ES and DE suggests that LON features also provide predictive information beyond the easy/hard distinction.
In contrast, prediction of \texttt{conv\_time} is notably less accurate ($R^2_{\text{cv}} \leq 0.65$), indicating that the present LON features do not sufficiently capture the landscape structures governing convergence speed.

\section{Limitations and Future Directions}
This study has the following limitations.
\begin{itemize}
    \item Our LON construction is specific to MSG landscapes. 
To apply our framework to benchmark sets such as BBOB, each benchmark function must first be approximated as an MSG landscape.
    \item The structures representable by MSG landscapes are constrained by the number of Gaussians, which limits both the diversity achievable by NS and the fidelity when approximating existing benchmark sets. 
    \item Our framework's parameters are fixed, with no sensitivity analysis performed. The optimizers considered are likewise limited to two evolutionary algorithms with fixed parameter settings and termination criteria.
\end{itemize} 

As future work, we plan to investigate the following research questions:
\begin{itemize}
    \item To what extent do BoA differences arising from different definitions or optimizers translate into differences in LON features?
    \item Can higher feature-space coverage of a problem set improve performance prediction accuracy, and if so, how much coverage suffices?
    \item How robust is the observed feature--performance link to other optimization algorithms and to the parameters of the MSG landscapes, LONs, and NS?
\end{itemize}

\section{Conclusions}
In this study, we proposed search-free LON construction and novelty-based landscape generation for studying the LON feature--performance link in continuous optimization.
Our framework exploits the structure of MSG landscapes to identify local optima analytically and to define their BoAs without iterative execution of a search algorithm.
The experimental results of RQ1 confirmed that the proposed BoAs closely align with gradient-based BoAs, supporting the validity of our search-free LON construction on MSG landscapes.
The experimental results of RQ2 showed that the proposed method increases feature-space coverage over random generation, and that the design of the initial population has a considerable impact on the resulting coverage.
Finally, the experimental results of RQ3 demonstrated that LON features of the generated instances correlate strongly with the success rates of CMA-ES and DE, enabling performance prediction with $R^2_{\text{cv}}$ ranging from $0.64$ to $0.94$.
These results indicate that our framework provides a systematic dataset of feature vector--performance pairs, which serves as a foundation for landscape-aware optimization.

\begin{credits}
\subsubsection{\ackname}
This work was supported by JSPS KAKENHI Grant Number JP25K21301 and by the Institute of Mathematics for Industry, Joint Usage/Research Center in Kyushu University. (FY2025 Short-term Joint Research “Optimization Problems with Diverse Structures Based on Quality-Diversity” (2025a045).)
\subsubsection{\discintname}
The authors have no competing interests to declare that are
relevant to the content of this article.
\end{credits}

\bibliographystyle{splncs04}
\bibliography{references}
\end{document}